\documentclass[letterpaper]{article} 
\usepackage{aaai25}  
\usepackage{times}  
\usepackage{helvet}  
\usepackage{courier}  
\usepackage[hyphens]{url}  
\usepackage{graphicx} 
\usepackage{natbib}  
\usepackage{caption} 
\usepackage{algorithm}
\usepackage{algorithmic}
\usepackage{multirow}
\usepackage{amsmath}
\usepackage{subfig}
\usepackage{booktabs}
\usepackage{amsfonts}
\usepackage{bm}

\usepackage{newfloat}
\usepackage{listings}
\DeclareCaptionStyle{ruled}{labelfont=normalfont,labelsep=colon,strut=off} 
\floatstyle{ruled}
\newfloat{listing}{tb}{lst}{}
\floatname{listing}{Listing}
\title{COSEE: Consistency-Oriented Signal-Based Early Exiting via Calibrated\\ Sample Weighting Mechanism}
\author{
    Jianing He\textsuperscript{\rm 1}, Qi Zhang\textsuperscript{\rm 1}, Hongyun Zhang\textsuperscript{\rm 1}, Xuanjing Huang\textsuperscript{\rm 2}, Usman Naseem\textsuperscript{\rm 3}, Duoqian Miao\textsuperscript{\rm 1}\thanks{Corresponding author}
}
\affiliations{
    \textsuperscript{\rm 1}Tongji University, China\\
    \textsuperscript{\rm 2}Fudan University, China\\
    \textsuperscript{\rm 3}Macquarie University, Australia\\

    \{jnhe, zhangqi\_cs, zhanghongyun, dqmiao\}@tongji.edu.cn, xjhuang@fudan.edu.cn, engr.usmannaseem87@gmail.com

}

\usepackage{bibentry}

\begin{document}

\maketitle

\begin{abstract}
Early exiting is an effective paradigm for improving the inference efficiency of pre-trained language models (PLMs) by dynamically adjusting the number of executed layers for each sample. However, in most existing works, easy and hard samples are treated equally by each classifier during training, which neglects the test-time early exiting behavior, leading to inconsistency between training and testing. Although some methods have tackled this issue under a fixed speed-up ratio, the challenge of flexibly adjusting the speed-up ratio while maintaining consistency between training and testing is still under-explored. To bridge the gap, we propose a novel Consistency-Oriented Signal-based Early Exiting (COSEE) framework, which leverages a calibrated sample weighting mechanism to enable each classifier to emphasize the samples that are more likely to exit at that classifier under various acceleration scenarios. Extensive experiments on the GLUE benchmark demonstrate the effectiveness of our COSEE across multiple exiting signals and backbones, yielding a better trade-off between performance and efficiency.
\end{abstract}

%

\section{Introduction}

Although tremendous improvements have been achieved by pre-trained language models (PLMs) in natural language processing tasks~\citep{bert,albert,GPT,roberta}, high computational costs of PLMs in both training and inference still hinder their deployment in resource-constrained devices and real-time scenarios. Besides, overthinking problem~\citep{Overthinking} also restricts the application of PLMs. Precisely, for easy samples, PLMs can generate correct predictions according to the representations indicated by shallow layers. However, high-level representations may focus on more intricate or unrelated details, leading to incorrect answers.

\begin{figure*}[!t]
  \centering
\includegraphics[width=0.75\textwidth]{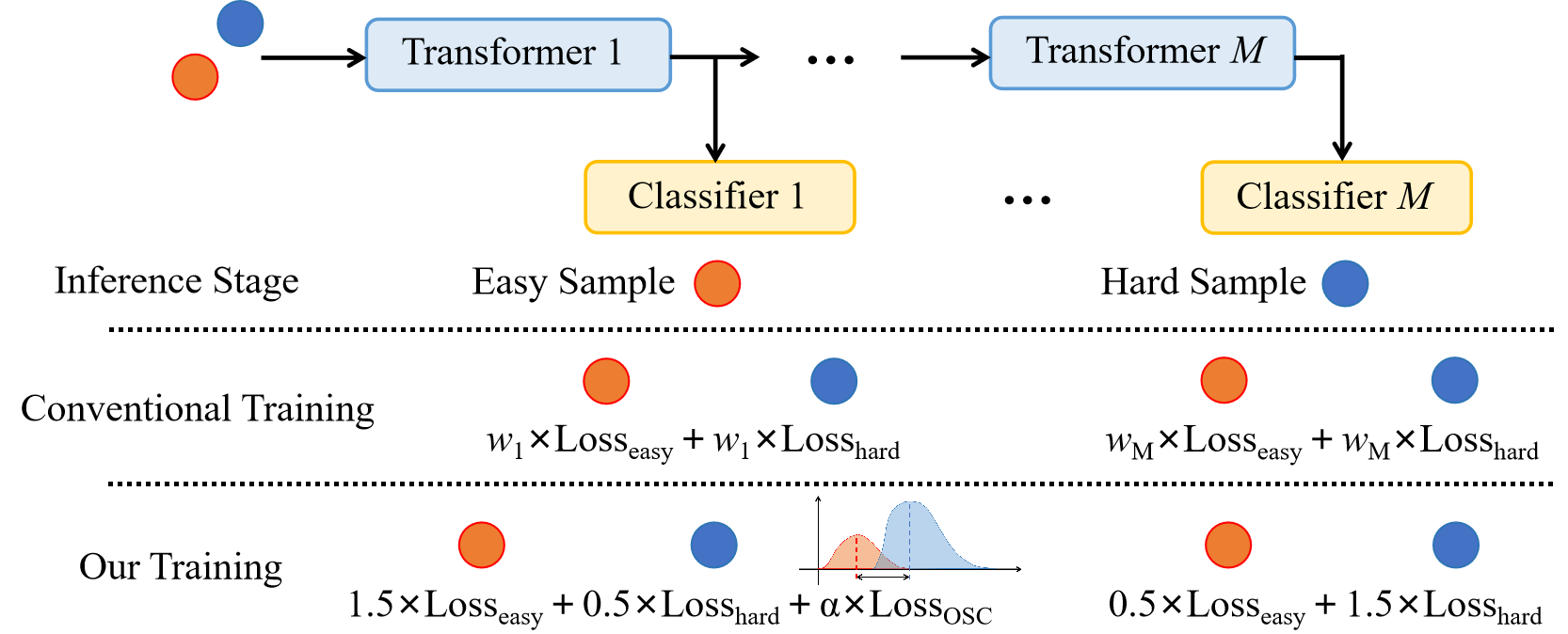}
  \caption{Comparison between the conventional signal-based early exiting framework and our COSEE. The conventional framework simply minimizes the (weighted) sum of cross-entropy losses from all classifiers, where each classifier treats all samples equally during training. Instead, our COSEE enables each classifier to emphasize samples that are more likely to exit at that classifier, ensuring consistency between training and testing. We also incorporate an online signal calibration objective $\rm Loss_{\rm OSC}$ for each internal classifier to encourage highly discriminative exiting signals for more reliable exiting decisions and loss weights.}
  \vspace{-2mm}
  \label{fig:overview}
\end{figure*}

To address these issues, early exiting~\citep{Deebert,Pabee,Berxit,globalpast,hashbased,ConsistentEE}, a kind of adaptive inference strategy, has been proposed to accelerate the inference of PLMs. As illustrated in Figure \ref{fig:overview}, each intermediate layer of the PLM is coupled with an internal classifier to give an early prediction. This enables the early exiting of samples once the early predictions are sufficiently reliable, eliminating the need for passing them through the entire model. This method employs a sample-wise inference strategy to deal with easy samples with shallow classifiers and process hard samples with deeper classifiers, significantly improving inference efficiency without sacrificing accuracy and alleviating the overthinking problem.

\begin{figure}[t]
 \centering
 \includegraphics[width=0.75\columnwidth]{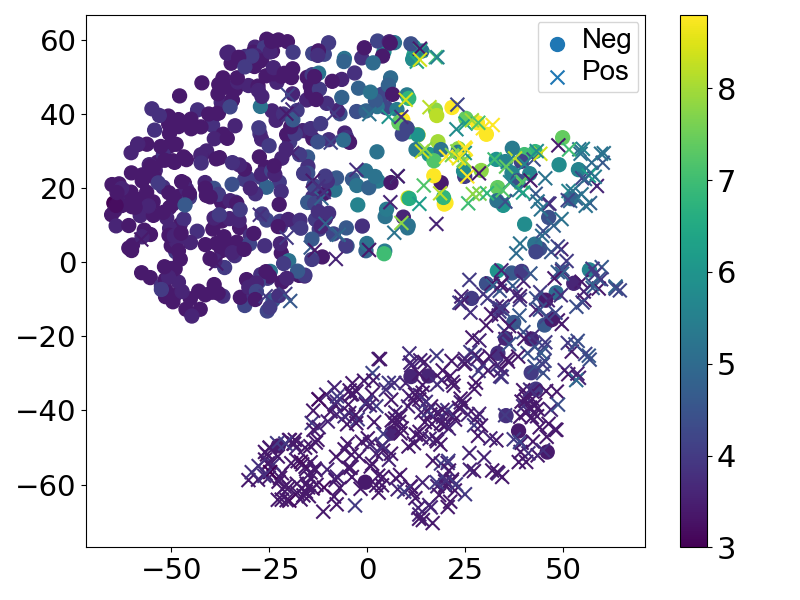}
  \caption{Exiting layer distribution on the QNLI development set with entropy-based exiting signal (Threshold = 0.4). Neg and Pos denote negative and positive samples, respectively. Samples near the classification boundary (hard samples) tend to exit at deep classifiers, while samples far from the classification boundary (easy samples) typically exit at shallow classifiers.  }
   \vspace{-2mm}
  \label{fig:intro}
\end{figure}

Signal-based early exiting methods~\citep{Deebert,Fastbert,Pabee,Righttool,cascadebert,globalpast,Disentangled,zhu2021leebert,GAML,F-PABEE,LECO,BADGE,ELANG,Berxit,Palbert} are typical implementations of early exiting, which rely on carefully designed exiting signals (e.g. entropy, energy score, softmax score, and patience) to dynamically adjust the number of executed layers for each sample. The inference process is terminated once the exiting signal meets a certain condition. These methods can easily adapt to various acceleration requirements during inference by simply adjusting the threshold, without incurring additional training costs. However, existing works simply use the (weighted) sum of cross-entropy losses from all classifiers as the training objective, where each classifier treats the loss of both easy and hard samples equally.
This treatment ignores the dynamic early exiting behavior during inference (as shown in Figure \ref{fig:intro}), leading to a gap between training and testing.

To bridge the gap, router-based early exiting methods~\citep{hashbased,BE3R,ConsistentEE} have been successively proposed. These methods employ a router (e.g. a hash function or a network) to determine the exiting layer of samples during both training and inference, and each sample only incurs a cross-entropy loss at its exiting classifier, ensuring consistency between training and testing. However, router-based early exiting methods fail to meet various acceleration requirements during inference, as a router can only generate a fixed exiting strategy, leading to unadjustable speed-up ratios.

In this paper, we aim to bridge the gap between training and testing while enabling flexible adjustments of the speed-up ratio. To this end, building upon the signal-based early exiting framework, we propose to assign sample-wise weights on the cross-entropy loss of all classifiers, such that each classifier is encouraged to emphasize samples that are more likely to exit at that classifier. Unfortunately, samples exit at different classifiers under various acceleration scenarios, bringing extreme challenges to weight assignment. 

To address the challenges, we propose a novel framework of Consistency-Oriented Signal-based Early Exiting (COSEE). Specifically, at each training step, we mimic the test-time early exiting process at multiple randomly selected thresholds to find where the samples tend to exit under different accelerations. Subsequently, we adopt a heuristic sample weighting mechanism (SWM) to assign weights on the cross-entropy loss of each sample across all classifiers, where each sample is emphasized by the classifiers near its exiting layer. Accordingly, we minimize the mean of cross-entropy losses across different thresholds to ensure the model's generalization ability in various acceleration scenarios. In addition, we further devise an online signal calibration (OSC) objective to generate highly discriminative exiting signals for more reliable exiting decisions, thus encouraging more proper loss weights based on exiting layers.

Our method is simple yet effective. Extensive experiments on the GLUE benchmark demonstrate that our COSEE framework with energy score consistently outperforms the state-of-the-art methods across all tasks, yielding a better trade-off between performance and efficiency with faster convergence speed and negligible additional storage overhead. In addition, an in-depth analysis further confirms the generalization of the COSEE framework on different exiting signals and backbones. Our main contributions can be summarized as follows:

\begin{itemize}
    \item We disclose that the performance bottleneck of current early exiting methods primarily stems from the challenge of ensuring consistency between training and testing while flexibly adjusting the speed-up ratios.
    \item We propose a novel Consistency-Oriented Signal-based Early Exiting (COSEE) framework to bridge the gap, which incorporates a sample weighting mechanism (SWM) and an online signal calibration (OSC) objective. 
    \item Extensive experiments verify the effectiveness of our COSEE across multiple exiting signals and backbones.
\end{itemize}

Code: \url{https://github.com/He-Jianing/COSEE}.

\section{Preliminaries}
In this section, we provide the necessary background for signal-based early exiting \footnote{Related works are detailed in Appendix \ref{sec:related_works}.}.

\subsection{Problem Definition}
Per Figure \ref{fig:overview}, given a BERT-style PLM with $M$ layers, we denote the hidden states at the $m$th layer as $h^{(m)}$. To enable early exiting during inference on a classification task involving $C$ classes, each intermediate layer is equipped with an internal classifier $F_m, m\in\{1, 2, \cdots, M-1\}$ to produce an early prediction $p^{(m)}=F_m(h^{(m)})$, i.e., a probability distribution over the $C$ classes. Classifiers in different layers do not share parameters.

\subsection{Signal-based Early Exiting}
For a given sample $x$, the inference process is terminated once the exiting signal at the current layer meets a certain condition. For exiting signals that exhibit a positive correlation with sample difficulty (e.g. entropy and energy score), early exiting is triggered once the exiting signal falls below a predefined threshold. A higher threshold leads to a higher speed-up ratio and potentially some performance degradation. Conversely, for exiting signals negatively correlated with sample difficulty (e.g. patience and softmax score),
the exiting condition is met when the exiting signal surpasses the threshold. A higher threshold leads to a lower speed-up ratio and performance improvements.  
 
\subsection{Conventional Training Methods}
In current signal-based early exiting methods, a widely used training objective involves the (weighted) sum of cross-entropy losses across all classifiers:
\begin{equation}\label{eq:conv_total_loss_sum}
L=\sum_{m=1}^M w_mL^{(m)},
\end{equation}
where $L^{(m)}$ denotes the cross-entropy loss of the $m$th classifier and $w_m$ denotes the corresponding loss weight. Under Eq.(\ref{eq:conv_total_loss_sum}), each classifier treats the loss of both easy and hard samples equally, which is inconsistent with the dynamic early exiting behavior during inference.

\section{The COSEE Framework}

\subsection{Framework Overview}
We propose a novel Consistency-Oriented Signal-based Early Exiting (COSEE) framework for PLMs, aiming to ensure consistency between training and testing while maintaining flexible adjustments of the speed-up ratio. Figure \ref{fig:overview} provides an overview of our framework. We first propose a sample weighting mechanism (SWM) that identifies the potential exiting layer of samples by simulating the test-time early exiting process during training and then uses this information to produce sample-wise loss weights across all classifiers. Additionally, we further devise an online signal calibration (OSC) objective to encourage highly discriminative exiting signals for more reliable exiting decisions, thus ensuring more proper loss weights based on exiting layers. Finally, regarding the exiting signal, we introduce a normalized energy score to align energy distributions across different layers for easy threshold selection. We primarily use it to implement the COSEE framework.

\subsection{Sample Weighting Mechanism}
Our goal is to identify the potential exiting layer of samples in various acceleration scenarios, and then assign greater weights to the cross-entropy loss of each sample on classifiers closer to its exiting layer. Accordingly, at each training step, all samples are passed through the entire model to generate predictions and exiting signals at all classifiers. Subsequently, we randomly select $K$ thresholds and simulate the early exiting process based on exiting signals at each threshold to find where the samples exit. This information is used to produce sample-wise loss weights across all classifiers.

{\bf Range for Threshold Selection.}  
For threshold selection, we collect the maximum and minimum values of exiting signals across all layers for training samples within each epoch and use them to create the selection range for the next epoch. We start with the thresholds randomly selected between 0 and 1 in the first epoch.

{\bf Weight Assignment.}
For a given threshold $\tau$, we impose sample-wise loss weights across all classifiers based on the exiting layer of samples and then compute the classification loss at threshold $\tau$:
\begin{gather}
L_{\rm CE,\tau}=\frac{1}{N}\sum_{n=1}^{N} \sum_{m=1}^{M} w_{n}^{(m)}\cdot \text{CE}(\hat{y}_{n}^{(m)},y_{n}), \label{eq:lce_tau} \\
w_{n}^{(m)}=\frac{e^{-\beta_{t}\cdot{\lvert m-m_{n}^{*}\rvert}}}{\sum_{m=1}^{M} e^{-\beta_{t}\cdot{\lvert m-m_{n}^{*}\rvert}}}, \label{eq:weight_n_m}
\end{gather}
where $\text{CE}(\hat{y}_{n}^{(m)},y_{n})$ and $w_{n}^{(m)}$ denote the cross-entropy loss and the loss weight for the $n$th sample at the $m$th classifier respectively, and $w_{n}^{(m)}$ satisfies $\sum_{m=1}^{M} w_{n}^{(m)}=1$. $N$ denotes the number of samples. $m_{n}^{*}$ denotes the index of exiting layer for the $n$th sample at threshold $\tau$, and $\beta_{t}$ denotes the decay factor at the $t$th training step. 
According to Equation~\ref{eq:weight_n_m}, classifiers closer to the exiting layer are assigned greater weights compared to those further away, i.e., each sample is emphasized by the classifiers near its exiting layer. Note that the loss weights of classifiers are symmetrical around the exiting layer for easy parameter selection. Different from router-based early exiting methods, which employ one-hot sample-wise loss weights such that each sample only incurs a cross-entropy loss on its exiting classifier, we employ a softer sample weighting mechanism to enable the generality of our COSEE on unseen thresholds.

During the early training stage, unstable exiting layers often lead to fluctuating loss weights, consequently impacting the model's convergence. To mitigate this problem, we conduct a warm-up operation for the decay factor $\beta_{t}$ to gradually increase the impact of the sample's exiting layers on the loss weights during training: 
\begin{equation}\label{eq:beta}
\beta_{t}=\gamma_{t}\cdot{\beta}_{0},  
\end{equation}
where ${\beta}_{0}$ is positive, and $\gamma_{t}$ is the ratio of the current training step to the total training steps.  
 
{\bf Classification Objective.}
To enable various acceleration ratios during inference, the classification objective is defined as the mean of classification losses across all $K$ thresholds:
\begin{equation}\label{eq:total_cls_loss}
L_{\rm CE}=\frac{1}{K}\sum_{\tau} L_{\rm CE,\tau}.
\end{equation}

\subsection{Online Signal Calibration}
While SWM effectively facilitates the training of multi-exit networks, exiting signals may not consistently reflect sample difficulty, particularly during the early training stages. This affects the reliability of exiting decisions, leading to sub-optimal loss weights based on exiting layers. Therefore, we introduce an online signal calibration (OSC) objective to explicitly enlarge the distribution divergence of exiting signals between easy and hard samples. Specifically, for exiting signals that indicate the sample difficulty (e.g. entropy and energy score), our OSC objective is formulated as:
\begin{gather}
L_{\rm OSC}=\frac{1}{M-1}\sum_{m=1}^{M-1} L_{\rm OSC}^{(m)}, \label{eq:total_osc_loss} \\
L_{\rm OSC}^{(m)}=\max(0, {\overline{S}_{easy}^{(m)}} - {\overline{S}_{hard}^{(m)}} + \epsilon), \label{eq:osc_m}
\end{gather}
where $L_{\rm OSC}^{(m)}$ is the signal calibration loss at the $m$th layer. $\overline{S}_{easy}^{(m)}$ and $\overline{S}_{hard}^{(m)}$ are the mean of exiting signals on easy and hard samples at the $m$th layer, respectively, and $\epsilon$ is the margin parameter shared across layers. For exiting signals negatively correlated with sample difficulty (e.g. softmax score), the calculation for $L_{\rm OSC}^{(m)}$ in Eq.(\ref{eq:osc_m}) needs to be replaced with:  
\begin{equation}\label{eq:osc_m_inv}
L_{\rm OSC}^{(m)}=\max(0, {\overline{S}_{hard}^{(m)}} - {\overline{S}_{easy}^{(m)}} + \epsilon).
\end{equation}

Note that we only minimize the signal calibration loss for the first $M-1$ layers, since there is no need to exit at the last layer. Additionally, we define samples as easy or hard depending on whether the internal classifier can predict them correctly, thus the partition may differ across layers.     
  
\subsection{Training Objective}
The training objective of the COSEE is formulated as the weighted sum of the classification and OSC objective:
\begin{equation}\label{eq:total_loss}
L=L_{\rm CE}+\alpha\times L_{\rm OSC},
\end{equation}
where $\alpha$ is a hyper-parameter used to balance the classification and OSC objectives. All internal classifiers are jointly trained with the backbone. 

\begin{figure}[!t]
\centering
\vspace{-4mm}
\subfloat[Original Energy Score] {\includegraphics[width=0.48\linewidth]{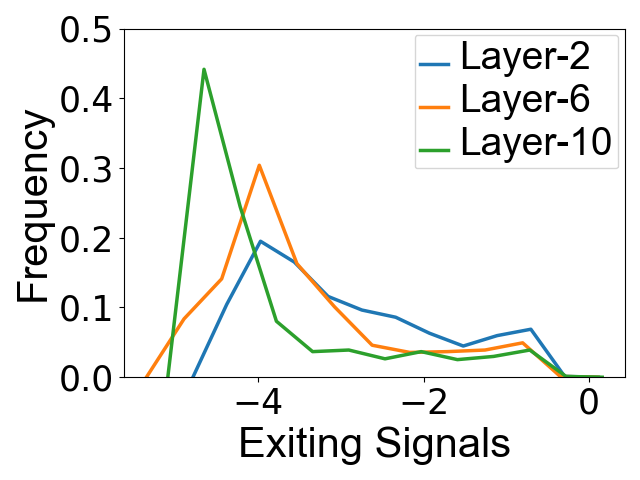}} 
\hfill
\subfloat[Normalized Energy Score] {\includegraphics[width=0.48\linewidth]{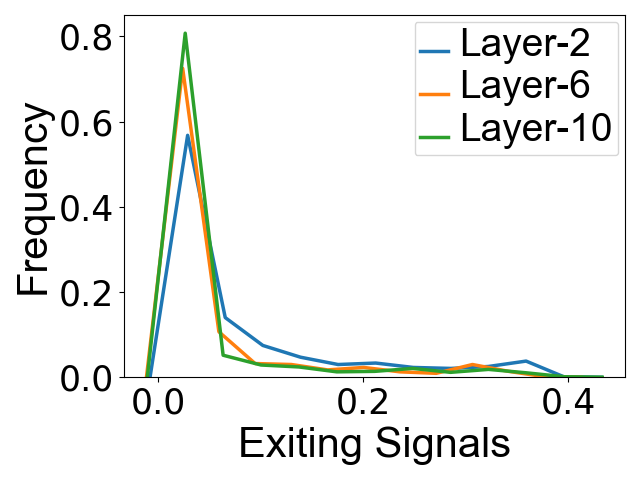}} 
\vspace{-1mm}
\caption {Energy distribution across layers 2, 6, and 10 for the SST-2 task. Normalization aligns the energy distribution across layers, facilitating threshold selection.}
\vspace{-2mm}
\label{fig:freq_distribution}
\end{figure}

\subsection{Exiting Signal}
Following E-LANG~\citep{ELANG}, we primarily implement our COSEE with the energy-based exiting signal. The energy score is defined as: 
\begin{equation}
\label{eq:energy_raw}
E(x;F_{m})=-\log \sum_{i=1}^{C} e^{f^{(m)}_{i}},
\end{equation}  
where $C$ is the number of classes, and $f^{(m)}_{i}$ denotes the logit value of sample $x$ on class $i$ suggested by the $m$th internal classifier $F_{m}$. A lower energy score indicates lower sample difficulty. The exiting criterion is met when the energy score falls below a predefined threshold. To align the energy distribution across different layers for threshold selection, we normalize the original energy scores to $(0,1)$:
\begin{equation}\label{eq:energy_norm}
E_{norm}(x;F_{m})={(1+e^{-E(x;F_{m})})}^{-1}.
\end{equation}
Figure \ref{fig:freq_distribution} 
confirms the superiority of the normalized energy score over the original energy score. In this paper, we mainly conduct experiments with the normalized energy score. Nevertheless, 
we also verify the effectiveness of the COSEE framework on other exiting signals, i.e., entropy and softmax score (see Section \ref{sec:generality}).    

\section{Experiments}
\label{sec:experiments}

\subsection{Tasks and Datasets}
Following \citet{cascadebert,globalpast}, we evaluate COSEE on six classification tasks from the GLUE benchmark~\citep{glue}, including SST-2, MRPC, QNLI, RTE, QQP, and MNLI. Data statistics are shown in Table~\ref{tab:data_statistics}.

\subsection{Baselines}
We compare our COSEE model with three groups of representative and state-of-the-art baselines.

{\bf Backbone.} We adopt the widely used BERT-base~\citep{bert} as the backbone for convincing comparisons.

{\bf Budget Exiting.}
We directly train a BERT-base with 6 layers (BERT-6L) to obtain a speed-up ratio of $2.00\times$, establishing a lower bound for early exiting methods as no techniques are employed.

{\bf Early Exiting.}
For signal-based early exiting methods, we choose DeeBERT~\citep{Deebert}, PABEE~\citep{Pabee}, BERxiT~\citep{Berxit}, LeeBERT~\citep{zhu2021leebert}, GPFEE~\citep{globalpast}, GAML-BERT~\citep{GAML}, PALBERT~\cite{Palbert}, and DisentangledEE~\citep{Disentangled}. For router-based early exiting methods, we choose state-of-the-art ConsistentEE~\citep{ConsistentEE}. Notably, some early exiting methods~\citep{hashbased,BE3R,LECO,BADGE} are not included due to the difference in backbones. CascadeBERT~\citep{cascadebert} and E-LANG~\citep{ELANG} are excluded for fair comparisons since they implement early exiting within several complete networks instead of a multi-exit network. Refer to Appendix~\ref{sec:baselines} for more details.

\begin{table}[!t]
  \centering
  \scalebox{0.8}{
  \begin{tabular}{lllll}
    \toprule
    Dataset & Classes &  $|$Train$|$  & $|$Test$|$ &  Task \\
    \midrule
    SST-2 & 2 &  67k &  1.8k & Sentiment \\
    MRPC & 2 &  3.7k &  1.7k & Paraphrase \\
    QQP & 2 &  364k &  391k & Paraphrase \\
    MNLI & 3 &  393k &  20k & NLI \\
    QNLI & 2 &  105k &  5.4k & QA/NLI \\
    RTE & 2 &  2.5k &  3k & NLI \\
    \bottomrule
  \end{tabular}}
  \caption{\label{tab:data_statistics}
    Dataset Statistics. NLI is the Natural Language Inference task, and QA is the Question Answering task.
  }
\end{table}

\begin{table*}[!t]
  \centering
  \vspace{-2mm}
  \scalebox{0.8}{
  \begin{tabular}{lllllll}
  \toprule
       \multirow{2}{1.0cm}{Method}   &  RTE  & MRPC  &  QQP  &  SST-2  &  QNLI  &  MNLI  \\
    &  \text{Acc} & \text{F1/Acc/Mean}&  \text{F1/Acc/Mean}& \text{Acc}& \text{Acc}& \text{Acc}   \\
    \midrule
    BERT-base & 66.4 (1.00$\times$) &  88.9/-/- (1.00$\times$)  & 71.2/-/- (1.00$\times$) &  93.5 (1.00$\times$) & 90.5 (1.00$\times$) & 84.6 (1.00$\times$) \\
    \midrule
    BERT-6L$^\dag$ & 63.9 (2.00$\times$) &  85.1/78.6/81.9 (2.00$\times$)  & 69.7/88.3/79.0 (2.00$\times$) &  91.0 (2.00$\times$) & 86.7 (2.00$\times$) & 80.8 (2.00$\times$) \\
    DeeBERT$^\dag$ & 64.3 (1.95$\times$) &  84.4/77.4/80.9 (2.07$\times$)  & 70.4/88.8/79.6 (2.13$\times$) &  90.2 (2.00$\times$) & 85.6 (2.09$\times$) & 74.4 (1.87$\times$) \\
     PABEE$^\dag$ & 64.0 (1.81$\times$) &  84.4/77.4/80.9 (2.01$\times$)  & 70.4/88.6/79.5 (2.09$\times$) &  89.3 (1.95$\times$) & 88.0 (1.87$\times$) & 79.8 (2.07$\times$) \\
     BERxiT & 65.7 (2.17$\times$) &  86.2/-/- (2.27$\times$)  & 70.5/-/- (2.27$\times$) &  91.6 (2.86$\times$) & 89.6 (1.72$\times$) & 82.1 (2.33$\times$) \\
     LeeBERT & - &  87.1/-/- (1.97$\times$)  & - &  92.6 (1.97$\times$) & - & 83.1 (1.97$\times$) \\
     GPFEE & 64.5 (2.04$\times$) &  87.0/81.8/84.4 (1.98$\times$)  &71.2/89.4/80.3 (2.18$\times$) &  92.8 (2.02$\times$) & 89.8 (1.97$\times$) & 83.3 (1.96$\times$) \\
     GAML-BERT & 64.3 (1.96$\times$) &  87.2/-/- (1.96$\times$)  & 70.9/-/- (1.96$\times$) &  92.8 (1.96$\times$) & 84.2 (1.96$\times$) & 83.3 (1.96$\times$) \\
     PALBERT$^\ddag$ & 64.3 (1.48$\times$) &  -/-/80.7 (1.48$\times$)  &-/-/79.3 (1.48$\times$) &  91.8 (1.48$\times$) & 89.1 (1.48$\times$) & 83.0 (1.48$\times$) \\
     DisentangledEE & 66.8 (1.25$\times$) &  -/-/83.8 (1.25$\times$)  &-/-/79.4 (1.25$\times$) &  92.9 (1.25$\times$) & 88.5 (1.25$\times$) & 83.0 (1.25$\times$) \\
     ConsistentEE & \bf{69.0 (1.85$\times$)} &  \bf{89.0/-/- (1.59$\times$)}  & -/89.0/- (1.82$\times$) &  92.9 (1.85$\times$) & 89.9 (1.72$\times$) & 83.4 (1.45$\times$) \\
     COSEE (ours) & 68.7 (1.96$\times$) &  88.0/82.0/85.0 (2.70$\times$)  & \bf{71.4/89.4/80.4 (2.01$\times$)} &  \bf{93.0 (2.14$\times$)} & \bf{90.2 (2.56$\times$)} & \bf{83.4 (1.92$\times$)} \\
     
    \bottomrule
   
  \end{tabular}}
  \caption{\label{tab:main_results}
     Performance comparison on the GLUE test set with BERT-base as the backbone. $\dag$ denotes the results taken from GPFEE~\citep{globalpast}, and $\ddag$ denotes the results taken from DisentangledEE~\citep{Disentangled}. Other baseline results are from their original papers. Our COSEE uses the normalized energy score as the exiting signal. Best results are marked in bold.}
     \vspace{-1mm}
\end{table*} 

\subsection{Experimental Settings}

{\bf Measurement.} Since the runtime is unstable across different runs, following~\citet{PCEE} and \citet{globalpast}, we utilize the saved layers to measure the speed-up ratio:
\begin{equation}\label{speedup-ratio}
\text{Speed-up Ratio}=\frac{\sum_{m=1}^M M\times N^m}{\sum_{m=1}^M m\times N^m},
\end{equation}
where $M$ is the total number of layers and $N^m$ is the number of samples exiting from the $m$th layer. According to~\citet{Deebert}, this metric is proportional to actual runtime.

{\bf Training.} Our implementation is based on Hugging Face's Transformers~\citep{huggingface}. Each internal classifier consists of a single linear layer. We mainly implement our COSEE framework with the normalized energy score if not specified. We also conduct experiments with entropy and softmax scores for generalization analysis. Following the previous work~\citep{Pabee,PCEE,globalpast}, we perform a grid search over learning rates of \{1e-5, 2e-5, 3e-5, 5e-5\}, batch sizes of $\{16, 32, 128\}$, $\alpha$ values in Eq.(\ref{eq:total_loss}) of $\{0.001, 0.01, 0.1, 1.0\}$, and $\beta_0$ values in Eq.(\ref{eq:beta}) of $\{0.05,0.2,1.0,10.0\}$. We set $\epsilon$ to $0.3$ in Eq.(\ref{eq:osc_m}) and $K$ to $5$ in Eq.(\ref{eq:total_cls_loss}). The maximum sequence length is fixed at $128$. We employ a linear decay learning rate scheduler and the AdamW optimizer~\citep{LoshchilovH19}. We conduct experiments on two RTX4090 GPUs with 24GB.

{\bf Inference.}
Following previous work~\citep{PCEE,globalpast}, we adopt a batch size of $1$ for inference, emulating a typical industry scenario where requests from various users arrive one by one. For fair comparisons, we carefully adjust the threshold $\tau$ for each task to achieve a similar speed-up ratio as the baseline methods (approximately $2.00\times$) and further compare the trade-off between task performance and inference efficiency.

\begin{figure}[!t]
\centering
\vspace{-3mm}
\subfloat[SST-2] {\includegraphics[width=0.50\linewidth]{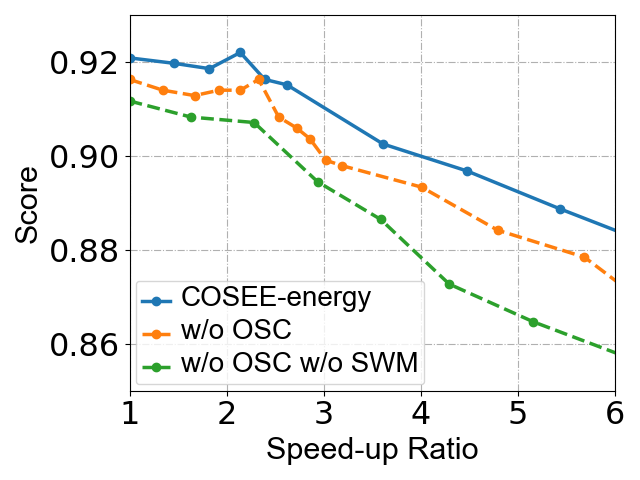}} 
\hfill
\subfloat[QNLI] {\includegraphics[width=0.50\linewidth]{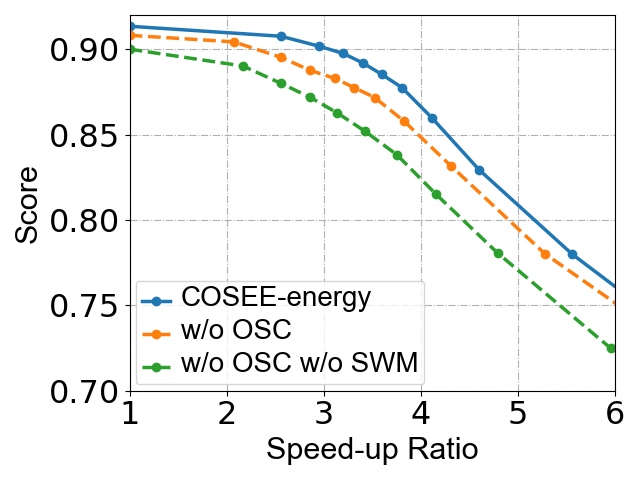}} 
\hfill
\vspace{-2mm}
\subfloat[MNLI] {\includegraphics[width=0.50\linewidth]{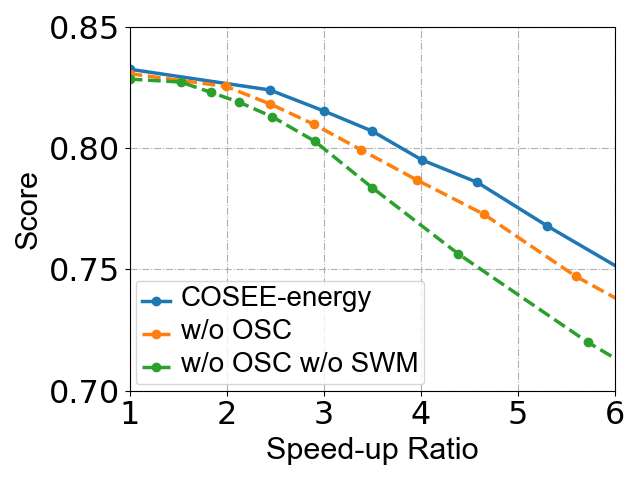}} 
\hfill
\subfloat[QQP] {\includegraphics[width=0.50\linewidth]{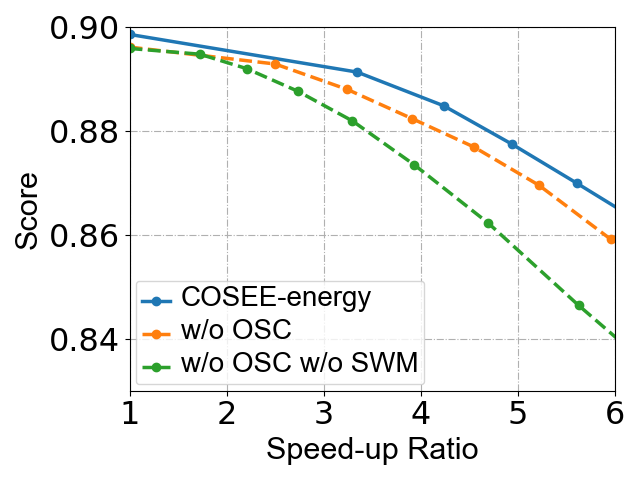}} 

  \caption {Impact of SWM and OSC on the trade-off between performance and efficiency for COSEE with energy on four GLUE development sets.}
  \label{fig:ablation}
  \vspace{-3mm}
\end{figure}

\subsection{Overall Performance Comparison}
Table \ref{tab:main_results} reports the test results of each early exiting method on the GLUE benchmark with BERT-base as the backbone model. The speed-up ratio is approximately $2.00\times$ (±38$\%$). Overall, our COSEE framework with normalized energy score demonstrates a superior performance-efficiency trade-off across different tasks compared to the baseline methods, which verifies the effectiveness of our design. Notably, our COSEE can even outperform the original BERT-base on RTE and QQP tasks, indicating that our method can effectively alleviate the overthinking problem of PLMs. This suggests that, for easy samples, predictions from intermediate layers may outperform those from the final layer. Our method enables easy samples to exit at shallow classifiers, thereby reducing the inference time while maintaining or even improving the task performance. Besides, our method can save training costs (see Section \ref{sec:ablation}) and introduce negligible additional storage overhead (see Section \ref{storage}).
We also explore the impact of hyperparameters (see Appendix~\ref{sec:pa}) and analyze the failure cases statistically (see Appendix~\ref{sec:failure}).

Although using energy scores and BERT-base in the primary experiments, we also verify the generality of COSEE on various exiting signals and backbones (see Section \ref{sec:generality}).

\subsection{Ablation Studies}
\label{sec:ablation}
{\bf Performance-Efficiency Trade-Off.}
To investigate the effectiveness of SWM and OSC, we plot the performance-efficiency trade-off curves of models trained using different methods on a representative subset of GLUE, as shown in Figure \ref{fig:ablation}. We can observe both SWM and OSC significantly improve the performance of early exiting across all tasks, especially under high speed-up ratios. This confirms the advantage of our COSEE under high acceleration scenarios, indicating the proposed SWM and OSC effectively facilitate the training of internal classifiers, particularly shallow ones.

\begin{figure}[!t]
\vspace{-0.4cm}
\centering
\subfloat[SST-2] {\includegraphics[width=0.50\linewidth]{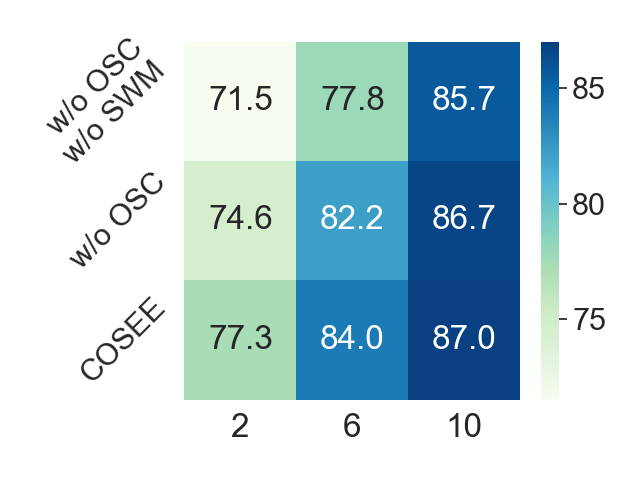}} 
\hfill
\subfloat[QNLI] {\includegraphics[width=0.50\linewidth]{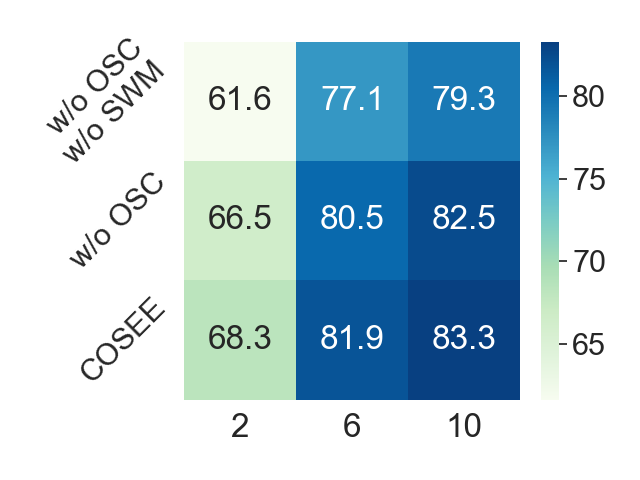}} 
  \caption {DIS heatmap of different models at different layers on the development sets of SST-2 and QNLI. Both SWM and OSC can encourage more discriminative exiting signals, further strengthening the reliability of exiting decisions.}
  \vspace{-2mm}
  \label{fig:dis_score}
\end{figure}

\begin{figure}[!t]
\vspace{-0.50cm}
\centering
\subfloat[SST-2] {\includegraphics[width=0.49\linewidth]{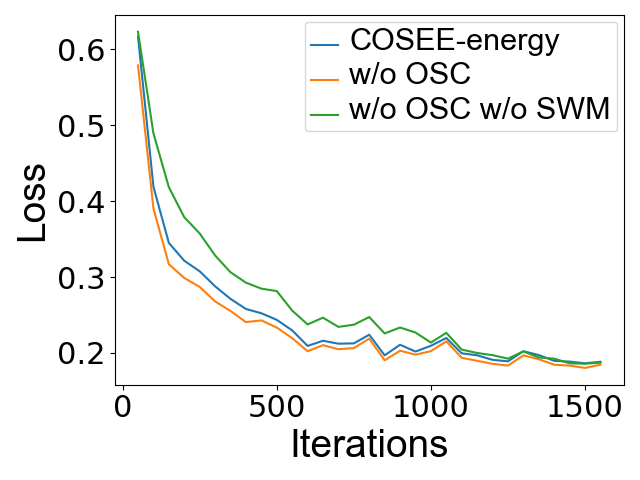}} 
\hfill
\subfloat[QNLI] {\includegraphics[width=0.49\linewidth]{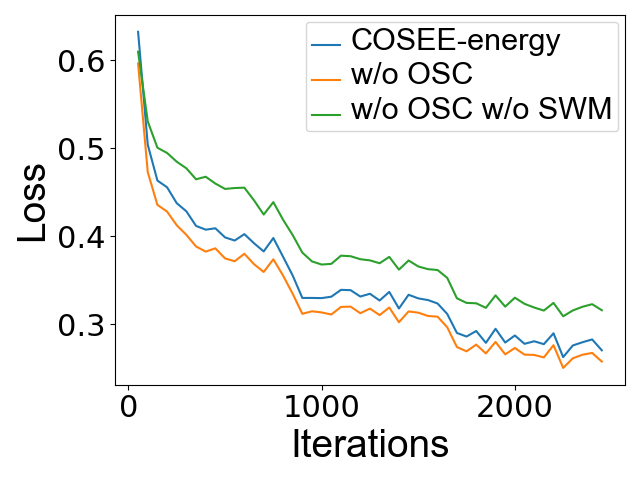}} 
\vspace{-1mm}
  \caption {Impact of SWM and OSC on training convergence for the SST-2 and QNLI tasks.}
  \label{fig:tcurve}
  \vspace{-2mm}
\end{figure}

{\bf Evaluation of Exiting Signals.}
Difficulty Inversion Score (DIS) is first proposed by~\citet{cascadebert}, an evaluation metric for exiting signals. A higher value indicates a greater correlation between the exiting signal and sample difficulty, thus enabling more reliable exiting decisions. Figure \ref{fig:dis_score} illustrates the DIS of exiting signals generated by different models on SST-2 and QNLI tasks. The results indicate that OSC explicitly enhances the correlation between the exiting signal and sample difficulty by enlarging the distribution divergence of exiting signals between easy and hard samples. Meanwhile, SWM encourages highly discriminative exiting signals by enabling each classifier to emphasize a subset of samples with certain difficulty levels. Also, it is noticeable that the improvements brought by SWM and OSC appear to be significant on shallow layers, which aligns with the observation shown in Figure \ref{fig:ablation}. This is due to the constrained capability of shallow classifiers, which allows for greater potential for improvements in training than deep classifiers.

{\bf Training Curves.}
To further explore the convergence speed of our COSEE during training, we plot the model's training curves across different training methods on SST-2 and QNLI tasks, as shown in Figure \ref{fig:tcurve}. The results indicate that the proposed SWM effectively accelerates the model's convergence during training. We attribute this to SWM's ability to reduce data complexity during training by enabling each classifier to emphasize a different subset of samples. Additionally, we can observe that incorporating the OSC objective can slightly impact the model's convergence speed. Nevertheless, our COSEE framework still maintains an advantage over the vanilla training method.

\begin{figure}[!t]
\vspace{-0.5cm}
\centering
\subfloat[train ($\tau=0.1$)] {\includegraphics[width=0.50\linewidth]{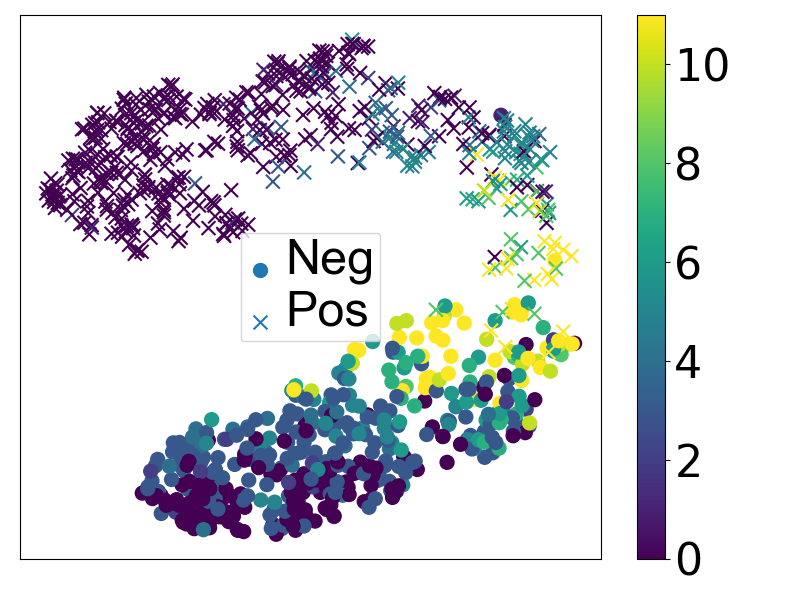}} 
\hfill
\subfloat[dev ($\tau=0.1$)] {\includegraphics[width=0.50\linewidth]{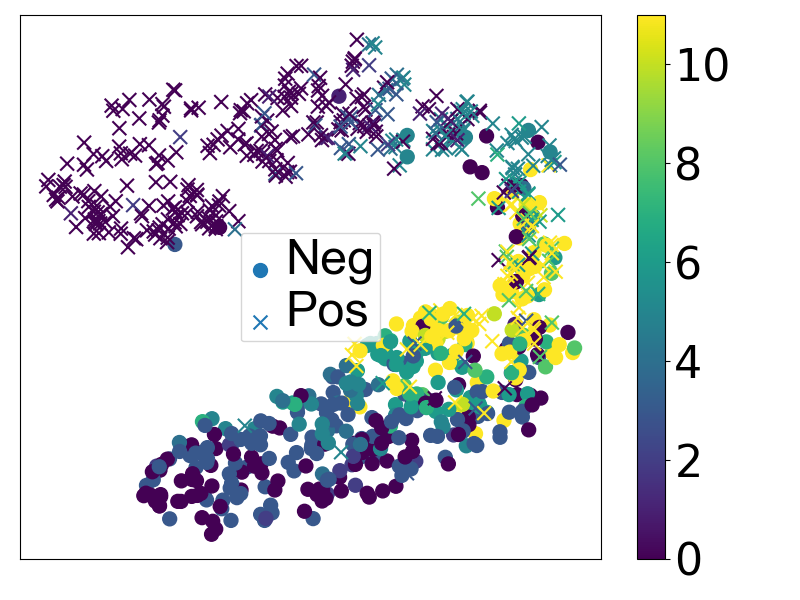}}
\vspace{-2mm}
\hfill
\subfloat[train ($\tau=0.2$)] {\includegraphics[width=0.50\linewidth]{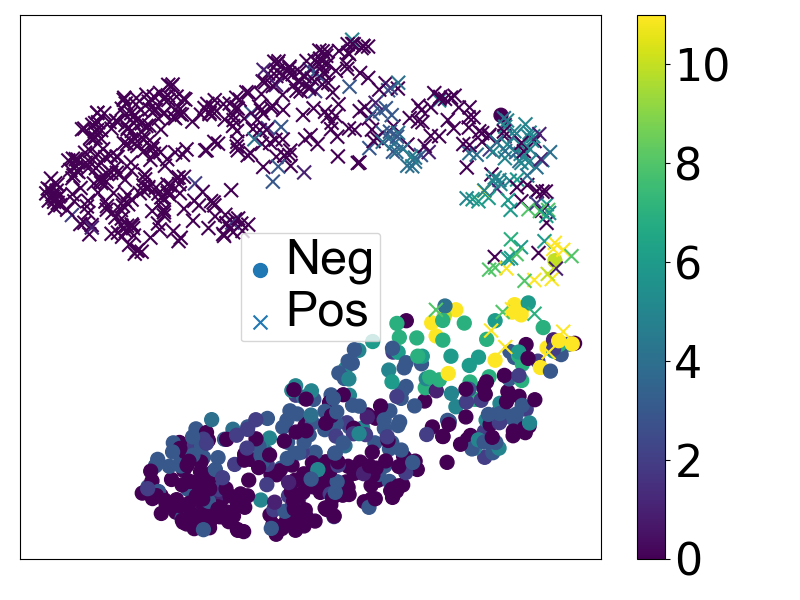}} 
\hfill
\subfloat[dev ($\tau=0.2$)]
{\includegraphics[width=0.50\linewidth]{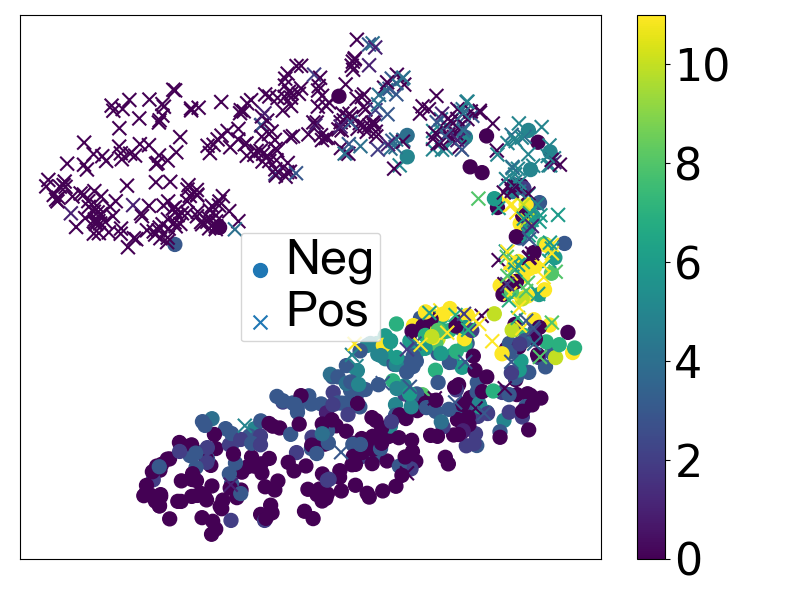}}
\hfill
\vspace{-2mm}
\subfloat[train ($\tau=0.3$)] {\includegraphics[width=0.50\linewidth]{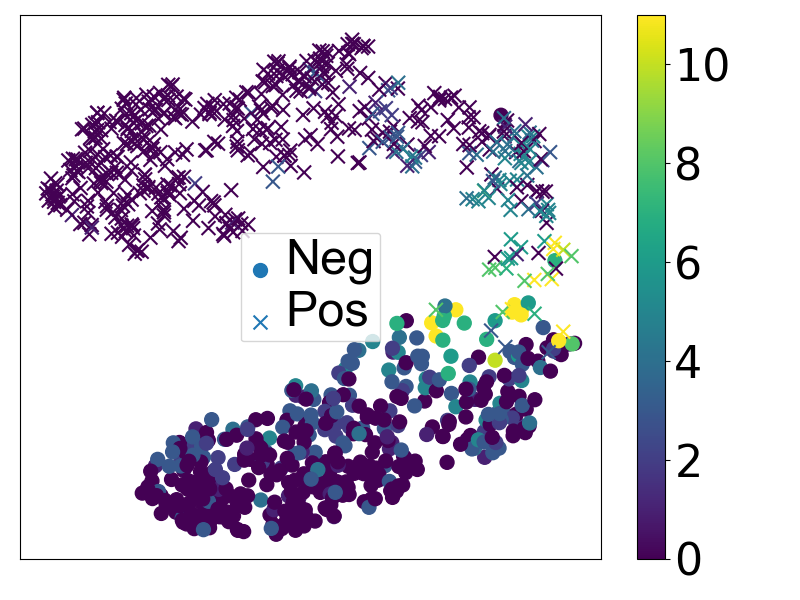}} 
\hfill
\subfloat[dev ($\tau=0.3$)]
{\includegraphics[width=0.50\linewidth]{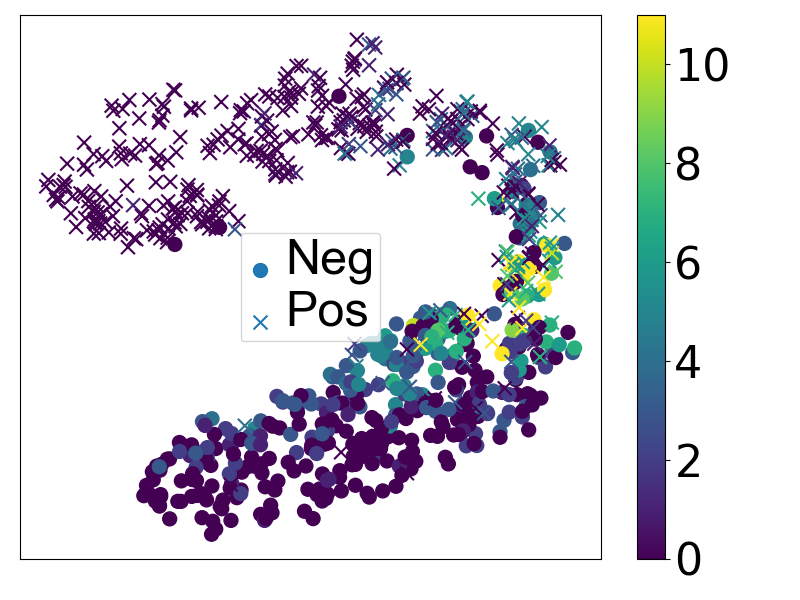}}
  \caption {Exiting layer distribution on the training and development sets of SST-2 task under different thresholds $\tau$. Neg and Pos denote negative and positive samples, respectively. The results exhibit consistency between training and development sets across different thresholds.}
  \vspace{-4mm}
  \label{fig:consistency}
\end{figure}

\section{In-depth Analysis}
\subsection{Visualization of Sample Exiting Layers}
To examine the consistency between training and testing under our COSEE framework, we visualize the exiting layer distribution in training and development sets at various thresholds, respectively. Figure \ref{fig:consistency} shows the visualization results for the SST-2 task. As we can see, training and development sets exhibit a consistent exiting layer distribution across different thresholds, which verifies the interpretability of our design. Additionally, we can observe that samples near the classification boundary (hard samples) tend to exit at deep classifiers while samples far from the classification boundary (easy samples) tend to exit at shallow classifiers. Furthermore, increasing the threshold will cause a decrease in exiting layers, thus achieving a higher speed-up ratio. These observations align with our intuitive understanding.

\begin{figure}[!t]
\centering
\vspace{-6mm}
\subfloat[SST-2] {\includegraphics[width=0.50\linewidth]{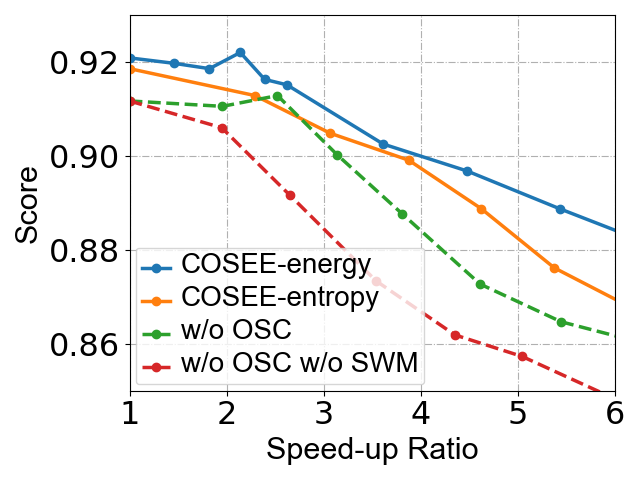}} 
\hfill
\subfloat[QNLI] {\includegraphics[width=0.50\linewidth]{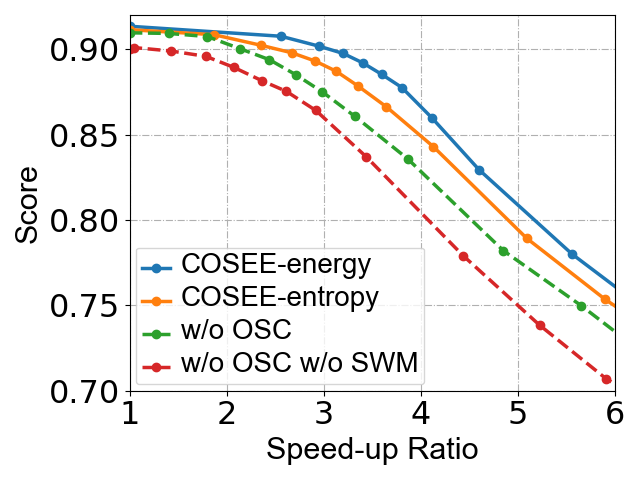}} 
\hfill
\vspace{-3mm}
\subfloat[MNLI] {\includegraphics[width=0.50\linewidth]{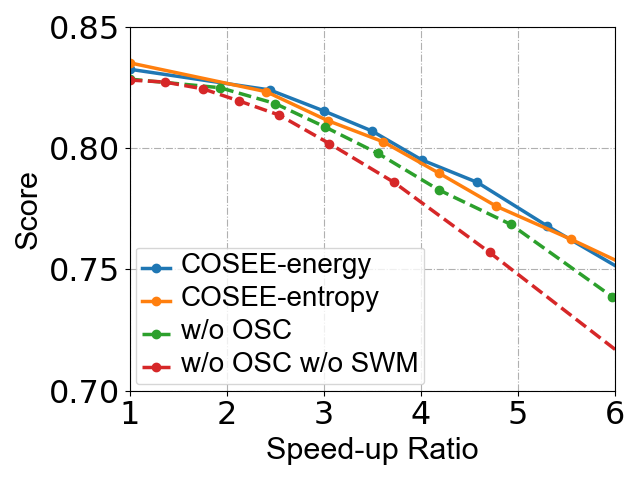}}
\hfill
\subfloat[QQP] {\includegraphics[width=0.50\linewidth]{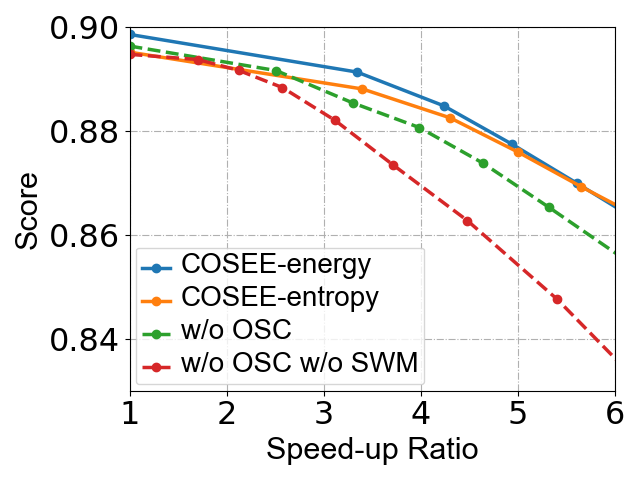}}

  \caption {Impact of SWM and OSC on the trade-off between performance and efficiency for COSEE with entropy. We include COSEE with energy in comparison. 
  }
  \label{fig:generality_entropy}
\end{figure}

\begin{figure}[!t]
\vspace{-4mm}
\centering
\subfloat[SST-2] {\includegraphics[width=0.50\linewidth]{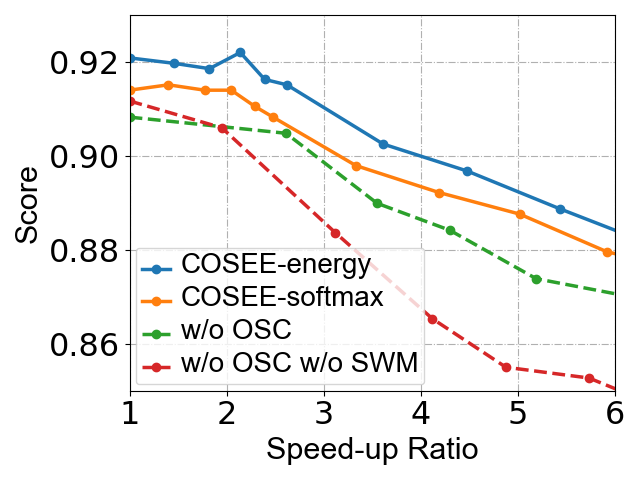}} 
\hfill
\subfloat[QNLI] {\includegraphics[width=0.50\linewidth]{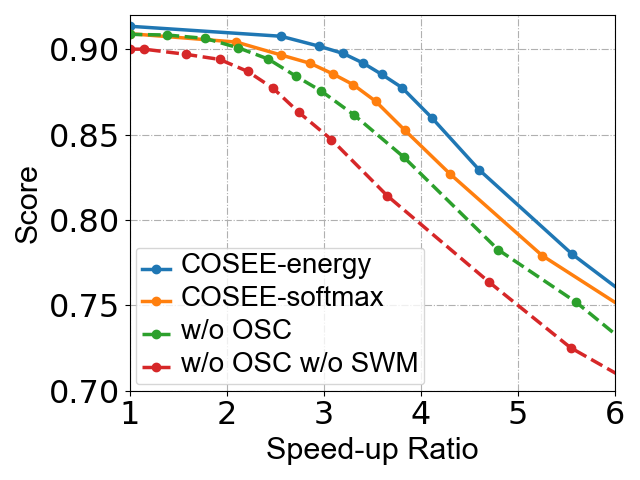}} 
\hfill
\vspace{-3mm}
\subfloat[MNLI] {\includegraphics[width=0.50\linewidth]{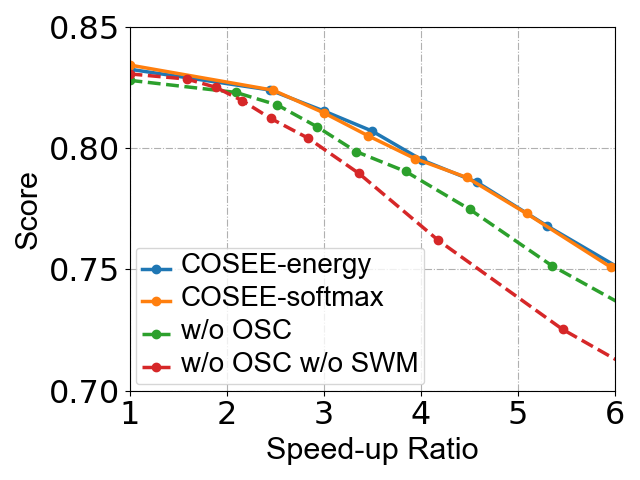}}
\hfill
\subfloat[QQP] {\includegraphics[width=0.50\linewidth]{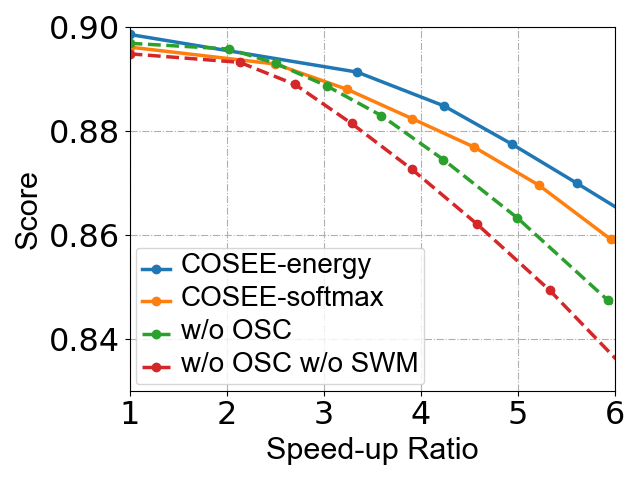}}

  \caption {Impact of SWM and OSC on the trade-off between performance and efficiency for COSEE with softmax score.
  }
  \vspace{-4mm}
  \label{fig:generality_softmax}
\end{figure}

\subsection{Generality of the COSEE Framework}
\label{sec:generality}
In this subsection, we explore the generality of our method on various exiting signals and backbones. The experiments are conducted on a representative subset of GLUE.

Figure \ref{fig:generality_entropy} and Figure \ref{fig:generality_softmax} present the experimental results of our COSEE framework with entropy and softmax scores, respectively. The results demonstrate the generality of our COSEE framework across various exiting signals. In addition, COSEE with energy scores outperforms COSEE with entropy or softmax scores on most tasks (SST-2, QNLI, and QQP). We speculate that this is because the energy score is more reliable in distinguishing easy and hard samples when compared to entropy and softmax scores. \citet{ELANG} confirmed this statement through theoretical derivation. Therefore, we primarily implement COSEE with energy scores in this paper. 

Table~\ref{tab:generality_albert} presents the performance comparison with backbone ALBERT-base. We observe that our COSEE with energy outperforms competitive baseline methods on most tasks, demonstrating its generality across different PLMs.  

\begin{table}
  \centering
  \vspace{-2mm}
  \scalebox{0.75}{
  \begin{tabular}{lllllll}
  \toprule
       Method &  Speed-up &  QQP  &  SST-2  &  QNLI  &  MNLI & AVG\\
  
    \midrule
    ALBERT-base$^\dag$ & 1.00$\times$ & 79.6  &  93.3  & 92.0  & 85.2  & 87.5 \\
    \midrule
    
    PABEE$^\dag$ & 1.95$\times$ &  \bf{79.8}  &  92.4  & 90.9  & 84.2  & 86.8 \\
    PALBERT & 1.21$\times$ &  79.1  &  91.4  & 90.9  & 83.2  & 86.2 \\
    DisentangledEE & 1.26$\times$ &  79.3  &  92.2  & 91.0  & 83.5  & 86.5 \\
     \bf{COSEE-energy}  & 2.12$\times$ &  79.6  &  \bf{92.9}  & \bf{91.8}  & \bf{84.8}  & \bf{87.3} \\

    \bottomrule
   
  \end{tabular}}
  \caption{\label{tab:generality_albert}
     Test results of different early exiting methods with ALBERT-base as the backbone. The speed-up ratio is averaged across 4 tasks. We report the mean of accuracy and F1-score for QQP, and accuracy for other tasks.
     $\dag$ denotes results taken from GPFEE~\citep{globalpast}. Other baseline results are taken from DisentangledEE~\cite{Disentangled}. } 
    \vspace{-2mm}
\end{table}

\subsection{Storage Costs Analysis}
\label{storage}
Table \ref{tab:storage_costs} compares the parameter volumes of our COSEE model with those of the original BERT-base. We observe that our COSEE model only requires less than 0.03$\%$ additional parameters due to the incorporation of internal classifiers. Additionally, it is noteworthy that the proposed SWM is parameter-free, yet it can effectively generate proper loss weights for each sample to facilitate the training of a multi-exit network.

\begin{table}[ht]
  \centering
  \scalebox{0.8}{
  \begin{tabular}{lcc}
\toprule
\multirow{2}{3.0cm}{Model} & \multicolumn{2}{c}{\#Params} \\
 & $C=2$ & $C=3$ \\
\midrule
BERT-base & 109.48M & 109.48M\\
COSEE & +16.92K & +25.38K\\
\bottomrule
\end{tabular}}
  \caption{\label{tab:storage_costs}
    Parameter volume Comparison. $C$ is the number of classes. COSEE introduces negligible extra parameters.  
  }
  \vspace{-2mm}
\end{table}

\section{Conclusion}
In this paper, we point out that the performance bottleneck of existing early exiting methods primarily lies in the challenge of ensuring consistency between training and testing while enabling flexible adjustments of the speed-up ratio. To remedy this, we propose COSEE, which mimics the test-time early exiting process under various acceleration scenarios based on calibrated exiting signals and then produces the sample-wise loss weights at all classifiers according to the sample's exiting layer. Our framework is both simple and intuitive. Extensive experiments on the GLUE benchmark demonstrate the superiority and generality of our framework across various exiting signals and backbones.

\clearpage
\section*{Acknowledgements}
This work was supported by the National Natural Science Foundation of China (Grant No. 62376198), the National Key Research and Development Program of China (Grant No. 2022YFB3104700), and the Shanghai Baiyulan Pujiang Project (No.~08002360429).

\bibliography{aaai25}

\clearpage
\appendix
\section{Parameter Sensitivity Analysis}
\label{sec:pa}
\begin{figure}[!t]
\centering
\subfloat[$\beta_{0}$-SST-2] {\includegraphics[width=0.50\linewidth]{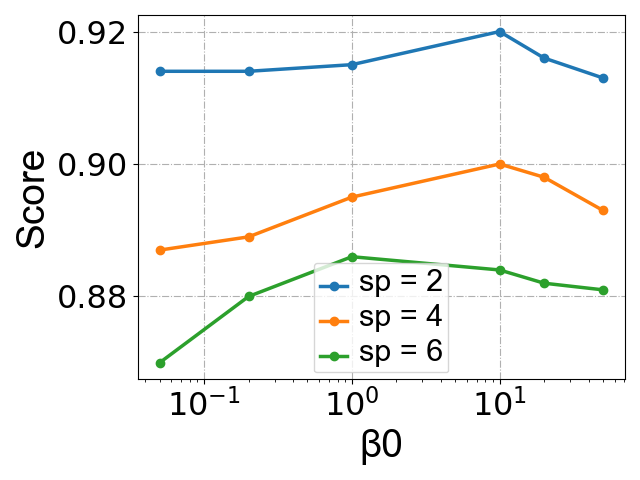}} 
\hfill
\subfloat[$\beta_{0}$-QNLI] {\includegraphics[width=0.50\linewidth]{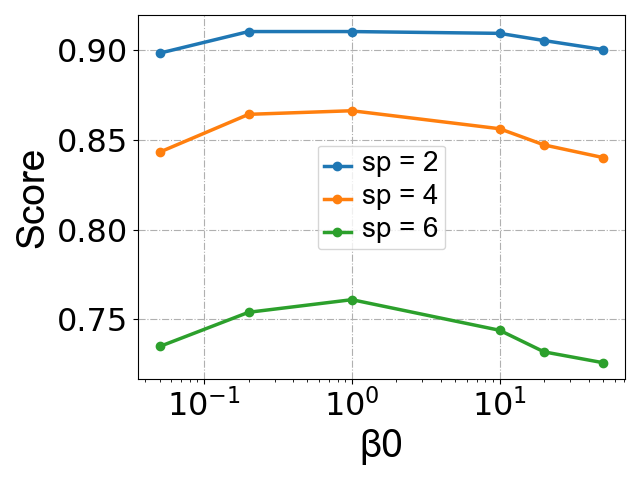}} 
\hfill
\vspace{-2mm}
\subfloat[$\alpha$-SST-2] {\includegraphics[width=0.50\linewidth]{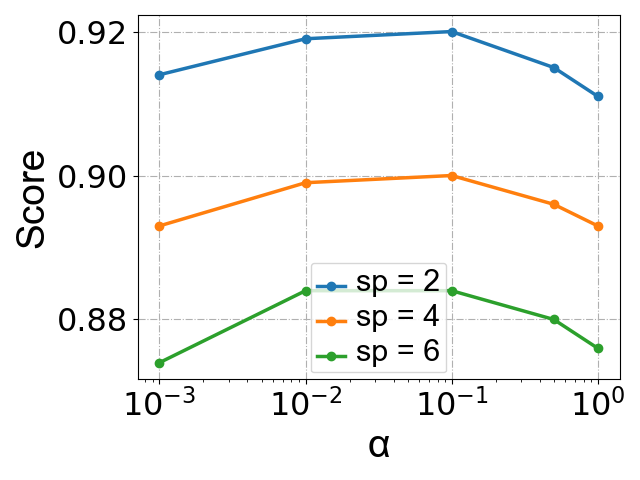}}
\hfill
\subfloat[$\alpha$-QNLI] {\includegraphics[width=0.50\linewidth]{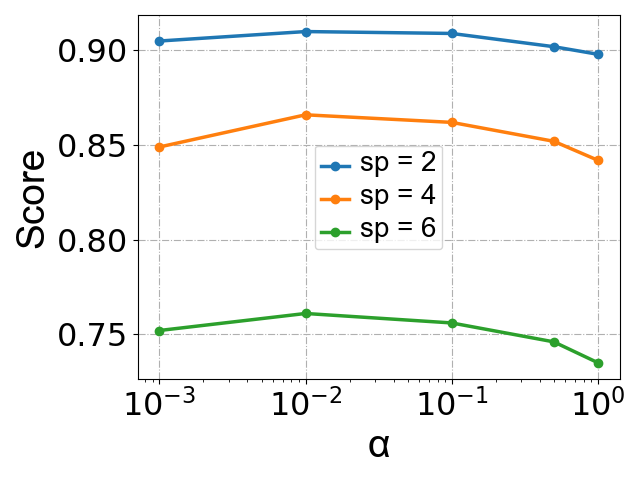}}
\hfill
\vspace{-2mm}
\subfloat[$\epsilon$-SST-2] {\includegraphics[width=0.50\linewidth]{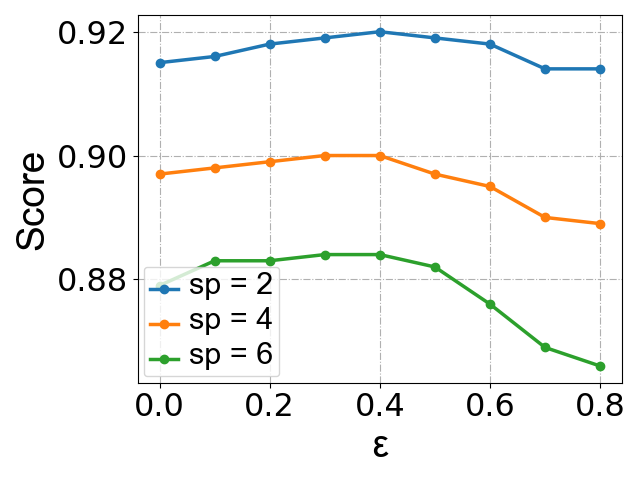}} 
\hfill
\subfloat[$\epsilon$-QNLI] {\includegraphics[width=0.50\linewidth]{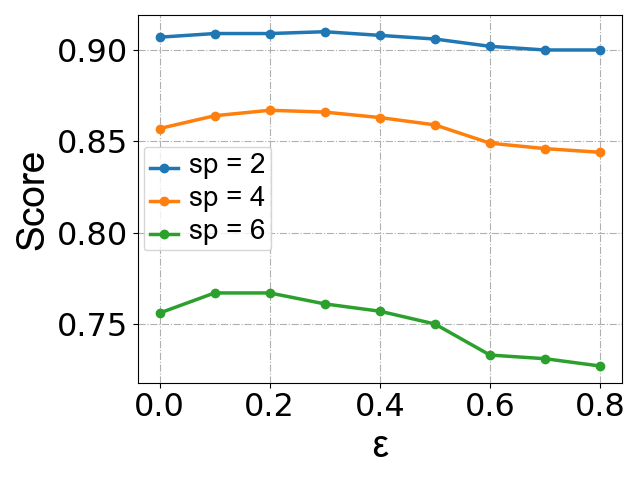}}
\hfill
\vspace{-2mm}
\subfloat[$K$-SST-2] {\includegraphics[width=0.50\linewidth]{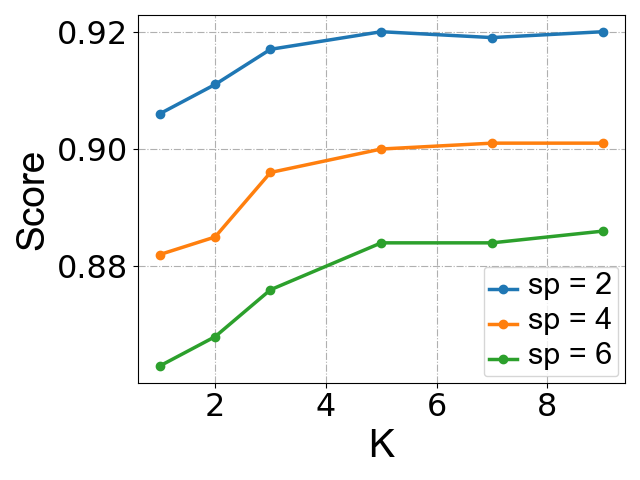}}
\hfill
\subfloat[$K$-QNLI] {\includegraphics[width=0.50\linewidth]{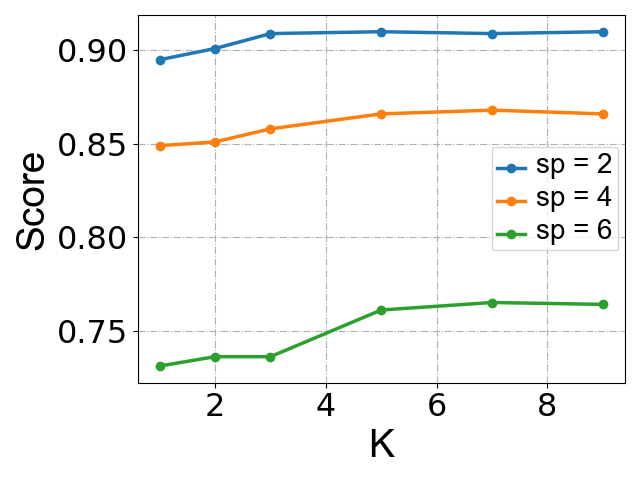}}
\vspace{-2mm}
  \caption {Sensitivity analysis of $\beta_0$, $\alpha$, $\epsilon$, and $K$ on the SST-2 and QNLI tasks. sp denotes the speed-up ratio. For each task, we start with a set of parameters ($\beta_0=10.0$, $\alpha=0.1$, $\epsilon=0.3$, $K=5$ for the SST-2 task, $\beta_0=1.0$, $\alpha=0.01$, $\epsilon=0.3$, $K=5$ for the QNLI task) and adjust the value of each parameter to generate curves, respectively.}
  \label{fig:pa}
  \vspace{-3mm}
\end{figure}

In this section, we further investigate the impact of four parameters on task performance under different speed-up ratios. Figure \ref{fig:pa} shows the experimental results on SST-2 and QNLI tasks.

{\bf Impact of $\beta_0$.}
The parameter $\beta_0$ in Eq.(\ref{eq:beta}) is utilized to adjust the distribution of loss weights across all classifiers. As shown in Figure \ref{fig:pa}(a) and Figure \ref{fig:pa}(b), both excessively large and small values of $\beta_0$ can impair the acceleration performance of the COSEE model, and the optimal $\beta_0$ value differs across tasks. Additionally, under high acceleration scenarios, the performance improvements brought by SWM are particularly significant, which is consistent with the observations in Figure \ref{fig:ablation} and Figure \ref{fig:dis_score}.

{\bf Impact of $\alpha$.}
The parameter $\alpha$ in Eq.(\ref{eq:total_loss}) balances the classification objective and OSC objective during training. The results depicted in Figure \ref{fig:pa}(c) and Figure \ref{fig:pa}(d) indicate that the acceleration performance of our COSEE model is not significantly affected by $\alpha$ selection, and an $\alpha$ value between 0.01 and 0.1 can always lead to satisfactory performance across different tasks. Moreover, an excessively large $\alpha$ value can lead to performance degradation. We attribute this to an overemphasis on the OSC objective, which interferes with the optimization of task performance.

{\bf Impact of $\epsilon$.}
The parameter $\epsilon$ in Eq.(\ref{eq:osc_m}) is utilized to regulate the distribution divergence of exiting signals between easy and hard samples. As shown in Figure \ref{fig:pa}(e) and Figure \ref{fig:pa}(f), an $\epsilon$ value between 0.1 and 0.4 can always lead to satisfactory performance across different tasks. Consequently, we fix the $\epsilon$ value at 0.3 to simplify the parameter selection. 

{\bf Impact of $K$.}
The parameter $K$ in Eq.(\ref{eq:total_cls_loss}) determines the number of thresholds randomly selected at each training step. The results in Figure \ref{fig:pa}(g) and Figure \ref{fig:pa}(h) show that, as the value of $K$ increases, the task performance exhibits a consistent pattern of initially increasing and then stabilizing under various speed-up ratios. This aligns with our intuition as selecting diverse thresholds during training is crucial for enabling models to meet different acceleration requirements during inference. We fix the $K$ value at 5 for all tasks for computational efficiency.

\section{Related Works}
\label{sec:related_works}
The early exiting methods applied to PLMs can be roughly divided into two categories: signal-based early exiting and router-based early exiting. 

\subsection{Signal-based Early Exiting}
Signal-based early exiting methods dynamically adjust the number of executed layers for each sample based on the exiting signal, enabling samples to exit early during inference once the exiting conditions are satisfied. 

According to the type of exiting signal, signal-based early exiting can be further categorized into score-based early exiting, patience-based early exiting, and learning-based early exiting. Score-based early exiting methods~\citep{Deebert,Fastbert,Righttool,ELANG,cascadebert,globalpast} leverage the entropy, softmax score, or energy score to capture the uncertainty of early predictions. Early exiting is triggered once the entropy or energy score (softmax score) falls below (surpasses) the predefined threshold. Patience-based early exiting methods~\citep{Pabee,F-PABEE,LECO,BADGE,Disentangled,zhu2021leebert,GAML} utilize the cross-layer consistency as the exiting signal. The exiting criterion is met when enough (i.e. greater than the threshold) consecutive internal classifiers agree with each other. Lastly, learning-based early exiting methods~\citep{Berxit,Palbert} leverage neural networks to generate exiting signals for exit decision-making.

According to the training of multi-exit networks, some studies~\citep{Deebert,Righttool,Berxit,Palbert} minimize the sum of cross-entropy losses across all classifiers, whereas other studies~\citep{Pabee,globalpast,zhu2021leebert,GAML} minimize the weighted sum of cross-entropy losses across all classifiers. In addition, LeeBERT~\citep{zhu2021leebert} and GAML-BERT~\citep{GAML} employ the cross-layer distillation objective to further enhance the training of internal classifiers.

Signal-based early exiting can easily adapt to different acceleration requirements by simply adjusting the threshold, without incurring additional training costs. However, in current studies, each classifier treats all samples equally during training, which neglects the dynamic early exiting behaviors across different samples, leading to a gap between training and testing.

\subsection{Router-based Early Exiting}
Router-based early exiting methods employ a router to determine exiting in both the training and inference phases. Some studies~\citep{hashbased,BE3R} utilize a hash function or a network to assign the exiting layer for each sample. ConsistentEE~\citep{ConsistentEE} employs reinforcement learning to train a policy network for exit decision-making. 

These methods perform early exiting during training, and each sample only incurs a cross-entropy loss at its exiting classifier. This treatment effectively ensures consistency between training and testing. However, router-based early exiting methods fail to meet various acceleration requirements during inference, as a router (a hash function or a trained network) can only generate a fixed exiting strategy, leading to unadjustable speed-up ratios.

This work proposes a novel signal-based early exiting framework to ensure consistency between training and testing while enabling flexible adjustment of the speed-up ratio.

\section{Baseline Competitors}
\label{sec:baselines}
In this section, we introduce the comparative early exiting methods in our experiments in detail. 

\subsection{Signal-based Early Exiting}
According to the exiting signal, DeeBERT and GPFEE leverage entropy to capture the uncertainty of early predictions. Early exiting is triggered once the entropy falls below the predefined threshold. PABEE, LeeBERT, GAML-BERT, and DisentangledEE employ the cross-layer consistency as the exiting signal. The exiting criterion is met when enough (i.e. greater than the threshold) consecutive internal classifiers agree with each other. BERxiT learns to score the correctness of early predictions through a network. The inference process is terminated when the correctness score surpasses the threshold.
PALBERT~\cite{Palbert} performs early exiting once the cumulative distribution function of the exiting layer’s probability distribution provided by neural networks exceeds the threshold.

According to the architecture or training objective of multi-exit networks, DeeBERT, PABEE, BERxiT and PALBERT minimize the cross-entropy losses for all classifiers. LeeBERT and GAML-BERT introduce the cross-layer distillation objective to encourage mutual learning among different classifiers. GPFEE integrates past and future states as inputs for each classifier, providing more reliable early predictions. 
DisentangledEE introduces an adapter to decouple the generic language representation learning and task-specific feature extraction and proposes a non-parametric simplex equiangular tight frame classifier for improvement.  

\subsection{Router-based Early Exiting}
ConsistentEE employs reinforcement learning to train a policy network for exit decision-making and minimizes the cross-entropy loss for each sample only at its exiting classifier during training.

\section{Statistics of Failure Cases}
\label{sec:failure}
In this section, we conduct a statistical analysis of failure cases to assess the reliability of exiting decisions.
There are two types of failure cases: first, the model may prematurely emit samples with incorrect early predictions by underestimating their difficulty, which degrades task performance; second, the model may delay the exiting of samples with correct early predictions by overestimating their difficulty, which prolongs inference time. Both types can negatively impact the model's performance-efficiency trade-off.
Accordingly, we define two metrics to statistically analyze these two types of failure cases, respectively.
\begin{itemize}
    \item \textit{Premature Exiting Rate}: the ratio of "exit" decisions made when the internal classifier predicts incorrectly.

    \item \textit{Delayed Exiting Rate}: the ratio of "continue" decisions made when the internal classifier predicts correctly.
\end{itemize}

\begin{figure}[ht]
\vspace{-0.40cm}
\centering
\subfloat[Premature Exiting Rate] {\includegraphics[width=0.49\linewidth]{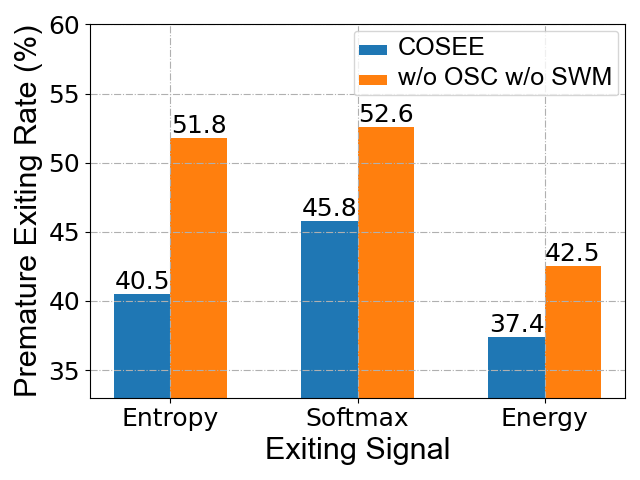}} 
\hfill
\subfloat[Delayed Exiting Rate] {\includegraphics[width=0.49\linewidth]{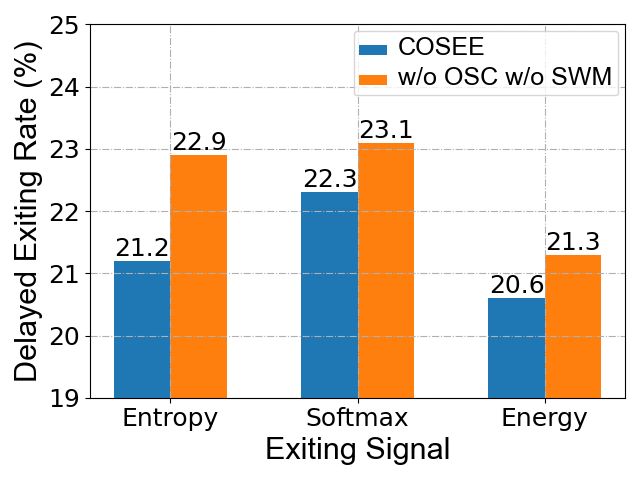}} 
\vspace{-1mm}
  \caption {Statistics of failure cases for each training method under various exiting signals. Results are based on the SST-2 development set at a $4.00\times$ speed-up ratio.}
  \label{fig:failure_cases}
  \vspace{-2mm}
\end{figure}

Exiting decisions are considered reliable only when both the Premature and Delayed Exiting Rate are sufficiently low.
Figure~\ref{fig:failure_cases} presents the failure cases statistics for COSEE and the conventional training method under various exiting signals in the SST-2 task. We observe that, compared to the conventional training method, our COSEE significantly reduces both the Premature and Delayed Exiting Rate across various exiting signals. This suggests that our COSEE facilitates the reliability of exiting decisions during inference by improving the training of multi-exit networks.

\end{document}